# On the Vulnerability of Fingerprint Verification Systems to Fake Fingerprints Attacks

J. Galbally-Herrero, J. Fierrez-Aguilar, J. D. Rodriguez-Gonzalez
F. Alonso-Fernandez, Javier Ortega-Garcia, M. Tapiador

Biometrics Research Lab.- ATVS, Escuela Politecnica Superior - Universidad Autonoma de Madrid
C/. Francisco Tomas y Valiente, 11 - Campus de Cantoblanco - 28049 Madrid, Spain
{javier.galbally, julian.fierrez, fernando.alonso, javier.ortega}@uam.es

## Abstract

*A new method to generate gummy fingers is presented. A medium-size fake fingerprint database is described and two different fingerprint verification systems are evaluated on it. Three different scenarios are considered in the experiments, namely: enrollment and test with real fingerprints, enrollment and test with fake fingerprints, and enrollment with real fingerprints and test with fake fingerprints. Results for an optical and a thermal sweeping sensors are given. Both systems are shown to be vulnerable to direct attacks.*

## 1  Introduction

In the current networked society there is an increasing need for reliable automated personal identification. Previous advances in pattern recognition have given rise to a technology area known as biometrics in which users are identified by what they *are* rather than by what they *know* (e.g. a password), which could be forgotten, or what they *have* (e.g. a key), which could be lost or stolen [1]. Within the field of biometrics, fingerprints are widely used in many personal identification systems due to its permanence and uniqueness [2]. The deployment of portable devices capable of capturing fingerprint images (e.g. mobile telephones or PC peripherals) is resulting in a growing demand of fingerprint-based authentication applications, such as access control or on-line identification.

However, in spite of their numerous advantages, biometric systems are also vulnerable to attacks, which can decrease their security [3]. In [3] eight different vulnerable points of a general biometric system are pointed out. All eight points are depicted in Fig. 1.

These attacks can be grouped in two generic classes: *i) direct attacks*, in which the intruder does not have any knowledge about the functioning of the system and which comprise attacks type 1 of Fig. 1, and *ii) indirect attacks* in which the impostor has some knowledge about the inner working of the system (e.g. the way data is stored) and which comprise all the remaining 7 attacks in Fig. 1.

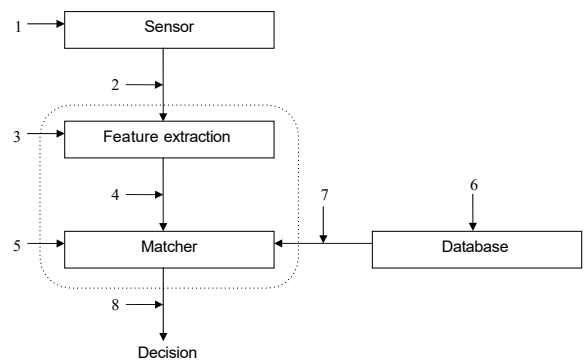

**Figure 1. Eight points of attack to a biometric verification system.**

In the present work we concentrate our efforts in studying direct attacks on fingerprint-based verification systems. For this porpuse we have generated a medium-size database with real fingerprints and their respective syntethic imitations on which two different fingerprint verification systems are tested, namely: the minutiae-based system of reference in fingerprint verification by NIST, and a ridge-based system developed in the ATVS group.

For this two systems evaluation is performed on three different scenarios: *i*) the normal operation mode in which both the enrollment and test are carried out using real fingerprints, *ii*) a second scenario where enrollment and test are done with fake fingerprints, and *iii*) a third scenario where the enrollment is performed with real fingerprints and the test with imitations. For all three cases results in terms of EER and DET curves are given and the robustness of both systems to direct attacks is analized.

This paper is structured as follows. Some related works are presented in Sect. 2. In Sect. 3 we detail the process followed for the creation of the gummy fingers, and the database used in the experiments is presented. The experimental protocol, some results and further discussion are reported in Sect. 4. Conclusions are finally drawn in Sect. 5.

## 2. Related Works

Putte and Keuning in [4] tested the vulnerability of several sensors to fake fingerprints made with plasticine and silicone. In this work two different methods to create fake fingerprints were described: with and without the cooperation of the user. When having the cooperation of the owner, the quality of the imitations increased and the system was easier to fool. Five out of the six sensors tested (optical and of solid state) accepted the gummy finger as real.

In [5] similar attacks to the ones described in [4] were carried out. In this case the fake fingerprints used were made of gelatin. Between 68 and 100% of the attempts for all the systems under study were fooled by the imitations.

Similar experiments testing different sensors and using several attacking methods can be found in [6] and [7].

## 3. Fake Fingerprints Database

A fast, reliable and fairly easy method to create gummy fingerprints from common materials is presented in the following sections. The process is divided into two steps: first the negative of the fingerprint is acquired (ridges and valleys in the fingerprint are respectively valleys and ridges in the negative), and then the gummy fingerprint is created from the negative.

### 3.1. Fake Fingerprint Generation Method

The material used for the negative is a special material sold in modeling shops commonly known as "modeling putty". This substance is a dual compound formed by two pastes. Once both pastes are mixed together we have several minutes before the mixture hardens and becomes solid. The steps followed in the creation of the negative are:

- Take a portion of both pastes and mix them uniformly. Spread the resulting ball on a sheet of paper and smooth the surface to eliminate non desired imperfections.
- Place the finger to be forged on the putty and press. Next lift the finger carefully and check that the fingerprint is thoroughly printed in the putty.

In Fig. 2 the negative of a fingerprint is shown.

Modeling silicone is used to create the final fake fingerprint. This silicone is sold in any modeling shop together with a catalyst that reacts with the silicone turning it into a consistent state. The steps followed in the creation of the gummy fingerprint are:

- Mix in a container the silicone with the catalyst.
- Pour the mixture over the negative of the fingerprint and wait until it hardens.
- Once the silicone hardens detach it from the negative.

In Fig. 3 an example gummy finger is shown.

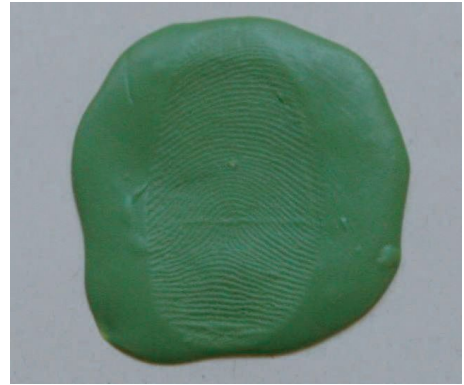

**Figure 2. Negative of the fingerprint.**

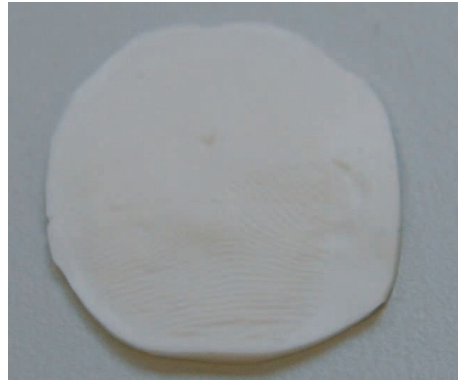

**Figure 3. Fake fingerprint.**

### 3.2. Database

Extending previous experiences in the acquisition of large scale databases [8], we generated a medium-size database of real fingerprints and their respective gummy imitations. The database was captured in just one session and it comprises the index and middle fingers of both hands of 17 donors, that is $4 \times 17 = 68$ genuine fingerprints and as many gummy imitations. Four images of each finger were acquired using two on-line fingerprint sensors, that is $68 \times 4 \times 2 = 544$ real samples and as many fake images. In order to accomplish a realistic acquisition and reach enough inter- and intra-variability, samples of the same finger were not acquired consecutively. The protocol followed for the acquisition of each user was:

- Index fingerprint / synthetic / right hand
- Middle fingerprint / synthetic / right hand
- Index fingerprint / synthetic / left hand
- Middle fingerprint / synthetic / left hand
- Index fingerprint / original / right hand
- Middle fingerprint / original / right hand
- Index fingerprint / original / left hand

- Middle fingerprint / original / left hand

As mentioned before two sensors were used in the acquisition of the database: an optical sensor and a thermal sweeping sensor. The protocol described above was repeated four times with each sensor in order to capture all the 4 impressions of each fingerprint.

### 3.2.1 Optical Sensor

The optical sensor used was the model Fx2000 by Biometrika, used in the Fingerprint Verification Competition 2002 (FVC 2002) [9]. It has a resolution of 569 dpi. In Fig. 4 we show some real fingerprints (top row) and their respective synthetic imitations (bottom row) captured with this device.

### 3.2.2 Thermal Sweeping Sensor

The thermal sensor used incorporates the fingerprint thermal sweeping technology by Atmel. This technology enables a very restricted sensor surface and a very compact architecture but it is more difficult to use and it may generate errors in the image reconstruction process. The resolution of the sensor is 500 dpi. In Fig. 5 we show some real fingerprints (top row) and their respective synthetic imitations (bottom row) captured with this device.

## 4 Experiments

### 4.1 Experimental Protocol

For the experiments each different finger in the database is considered as a different user. This way we have 4 real impressions and 4 synthetic impresions of all 68 users (17 donors and 4 fingers per donor). Three different scenarios are considered in the experiments:

- **Scenario 1**: both the enrollment and the test are carried out with real fingerprints. This corresponds to the normal operation mode of the system and is used as the scenario of reference.
- **Scenario 2**: both the enrollment and the test are carried out with fake fingerprints. In this case the attacker enrolls to the system with the fake fingerprint of the genuine user and then tries to access the application with that same fake fingerprint.

For the previous two scenarios the following sets of scores are generated: *i*) for genuine tests all the 4 samples of each user (real or fake depending on the scenario) are matched with each other avoiding symmetric matchings $((4 \times 3)/2 = 6$ scores per user), this results in $6 \times 68 = 408$ genuine scores, and *ii*) for impostor tests each of the four samples of every user are matched with all the samples of the remaining users in the database, resulting in $67 \times 4 \times 4 \times 68 = 72896$ impostor scores.

- **Scenario 3**: the enrollment is done using real fingerprints and tests are carried out with fake fingerprints. In this case the genuine user enrolls with his fingerprint and the attacker tries to access the application with the gummy fingerprint of the legal user.

In this last scenario the genuine set of scores is computed the same way as in scenario 1. Impostor scores are computed matching all 4 original samples of each user with all 4 fake samples which results in $16 \times 68 = 1088$ impostor scores.

This experimental protocol was followed for the two sensors used in the database acquisition so the sets of scores mentioned above were computed twice.

For each of the three scenarios described, the performance of two different fingerprint verification systems was evaluated:

- The minutiae-based NIST Fingerprint Image Software 2 (NFIS 2) [10].
- The ridge-based fingerprint verification system developed in the Biometrics Research Lab. ATVS (Univ. Autonoma de Madrid) [11, 12].

### 4.2 Results

#### 4.2.1 Quality Measures

The public software from NIST for fingerprint verification provides an application for the computation of quality in fingerprint images. This program ranks the samples in five levels of quality (from 1 to 5), being 1 the highest quality and 5 the lowest. In Fig. 6 we show the quality distributions of the database samples, both real and fake. The left plot depicts the distribution for samples acquired with the optical sensor and the right plot shows the distribution for the thermal sensor samples.

As we can observe in Fig. 6, the quality of the real samples captured with both sensors is very high, being most of the images concentrated in the first two levels. The quality of the real samples acquired with the optical sensor is higher due to the image reconstruction process carried out in the sweeping thermal technology. The quality of the fake fingerprints captured with the optical sensor is acceptable, most of the samples are concentrated in levels 2 and 3, but it is still clearly lower than the quality of the real fingerprints acquired with that same sensor. In the right plot of Fig. 6 we can observe a very significant decrease in the quality of the fake fingerprints acquired with the thermal sensor compared to the real ones.

The big difference of quality in the gummy imitations is due to the different technology used in both cases. The optical sensor is based on refraction effects of the light which take place in a similar way both in the skin and in the silicone, thus we obtain good quality imitations. On the other hand, the thermal sensor is based on the difference of temperature between ridges and valleys which is almost non-existent in the silicone fingerprints. Although the imitations were heated up breathing on them to get the necessary temperature difference, the quality of the samples is very poor.

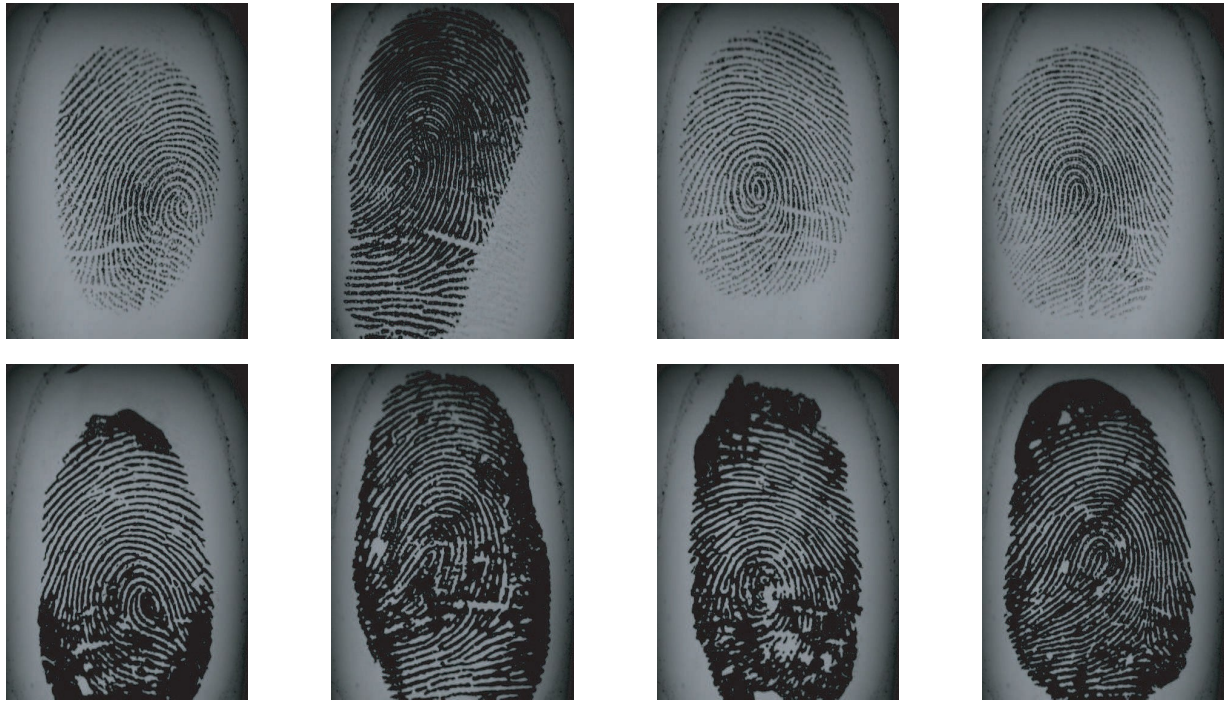

**Figure 4. Real samples captured with the optical sensor (top row) and their respective imitations (bottom row).**

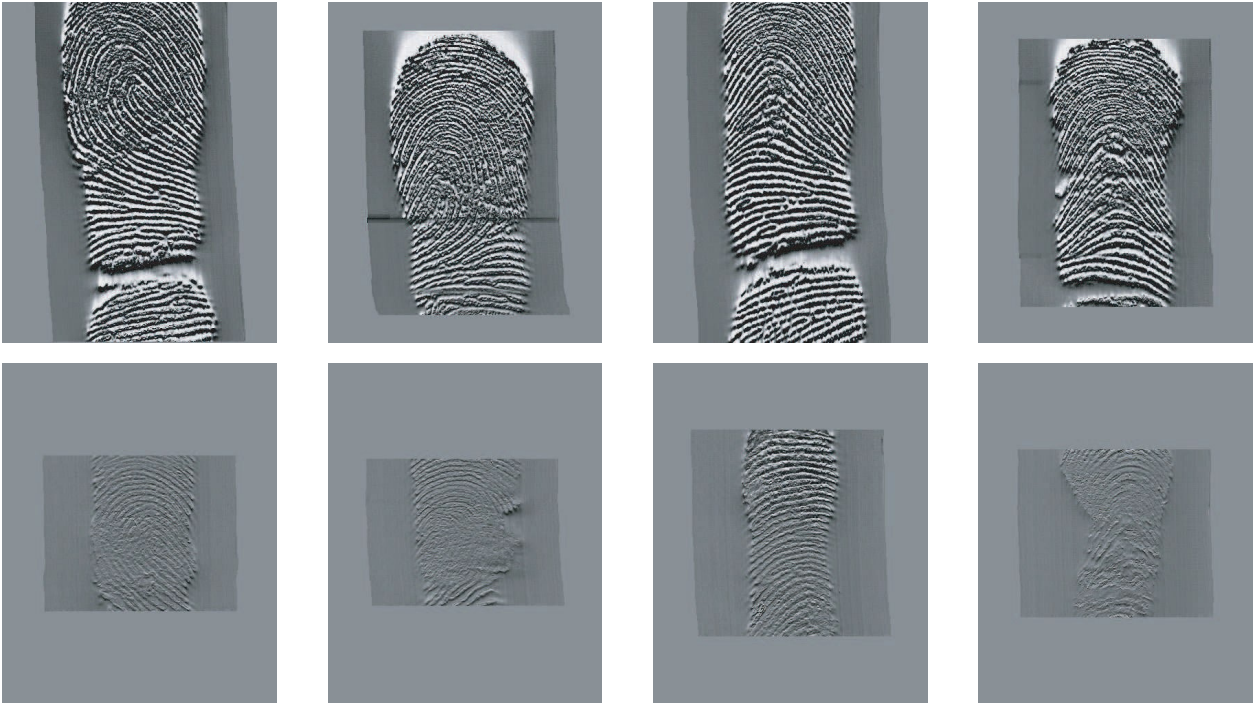

**Figure 5. Real samples captured with the thermal sweeping sensor (top row) and their respective imitations (bottom row).**

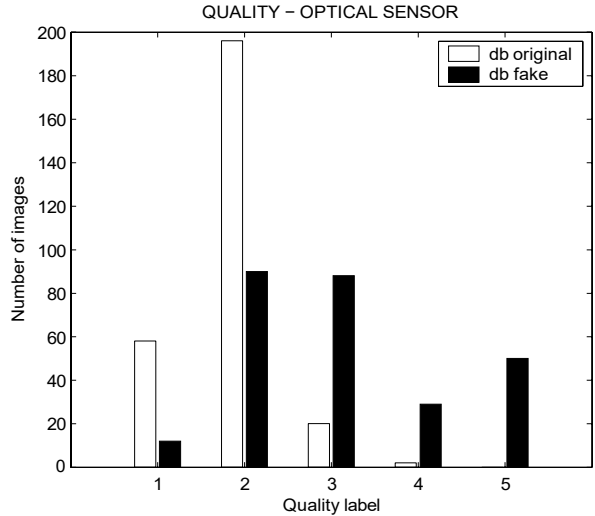 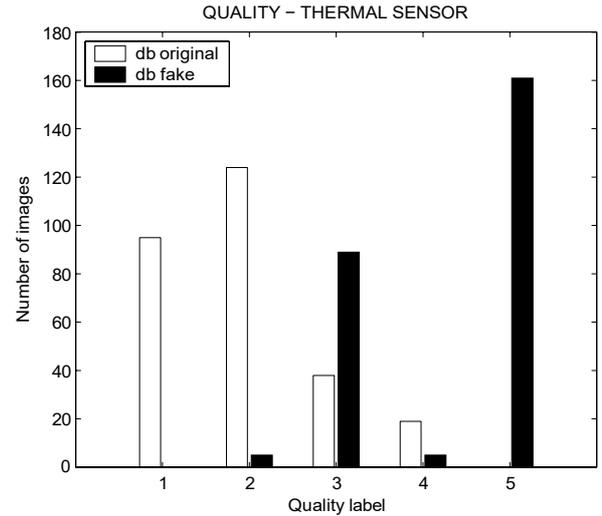

**Figure 6. Quality measures for the real and fake fingerprints acquired with the optical sensor (left plot) and the thermal sensor (right plot).**

**Table 1. EER for the NIST fingerprint verification system on the 3 scenarios considered using the optical and the thermal sensors.**

|  | EER (%) | | |
| --- | --- | --- | --- |
|  | Scenario 1 | Scenario 2 | Scenario 3 |
| Optical | 0.22 | 4.83 | 12.13 |
| Thermal | 6.42 | 39.46 | 7.60 |

**Table 2. EER for the ATVS fingerprint verification system on the 3 scenarios considered using the optical and the thermal sensors.**

|  | EER (%) | | |
| --- | --- | --- | --- |
|  | Scenario 1 | Scenario 2 | Scenario 3 |
| Optical | 11.27 | 15.05 | 22.55 |
| Thermal | 19.85 | 44.12 | 32.35 |

#### 4.2.2 Evaluation of the NIST System

In Table 1 we show the EER for the NIST fingerprint verification system, for the three evaluation scenarios considered and for the two sensors used. This values are extracted from the DET curves shown in Fig. 7, being the optical sensor depicted in the left plot and for the thermal sensor in the rigth plot.

**Optical sensor**. As expected from the quality measures, the performance of the system when enrolling and testing with real fingerprints (scenario 1) is very high, EER=0.22%. This performance drops when considering the second and third scenarios. In scenario 3 the EER is 12.13% which is much bigger than in the first scenario, showing a very high level of vulnerability of the system to direct attacks.

**Thermal sensor**. In Table 1 we can see that the EER of the system in the normal operation mode (scenario 1) with the thermal sensor is much higher (6.42%) than with the optical one (0.22%). Due to the very low quality of the fake samples captured with the thermal sensor, the EER hardly varies from the first scenario (6.42%) to the third (7.40%), which implies that fake samples acquired with the thermal sensor have very little valid information. Furthermore, in the scenario 2 the EER of the system is 39.46%, which corroborates the fact that fake samples of the thermal sensor have no discriminative power on the NIST system.

In summary, the NIST fingerprint verification system performs better with the optical sensor but it is also more vulnerable to direct attacks on it as compared to experiments on the thermal sensor.

#### 4.2.3 Evaluation of the ATVS System

In Table 2 we show the EER for the ATVS fingerprint verification system, for the three evaluation scenarios considered and for the two sensors used. In Fig. 8 the curves from which these EER values were obtained are depicted. In the left plot the curves for the optical sensor are shown and in the right plot for the thermal sensor.

**Optical sensor**. As expected from a ridge-based matcher its performance in the normal operation mode is lower than that of the minutiae-based system. However, there is just a factor 2 increase in the EER from scenario 1 to scenario 3, compared to more than an order of magnitude rise of the NIST matcher. Thus, the ridge-based system is less vulnerable to direct attacks than the minutiae-based one.

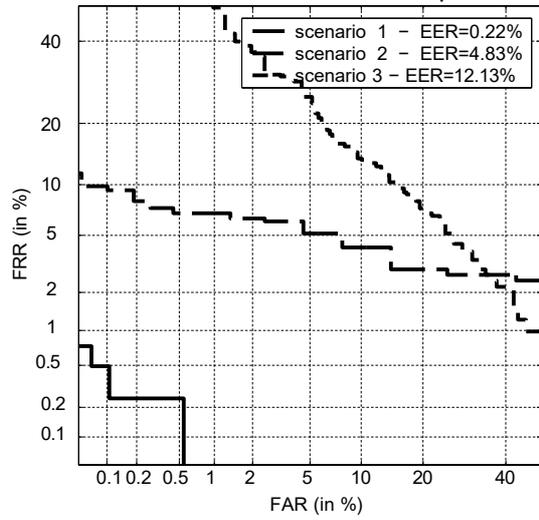 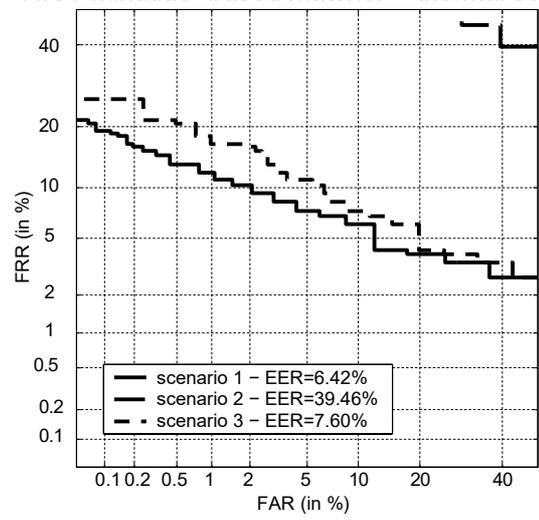

**Figure 7. DET curves for the NIST system using the optical sensor (left) and the thermal sensor (right).**

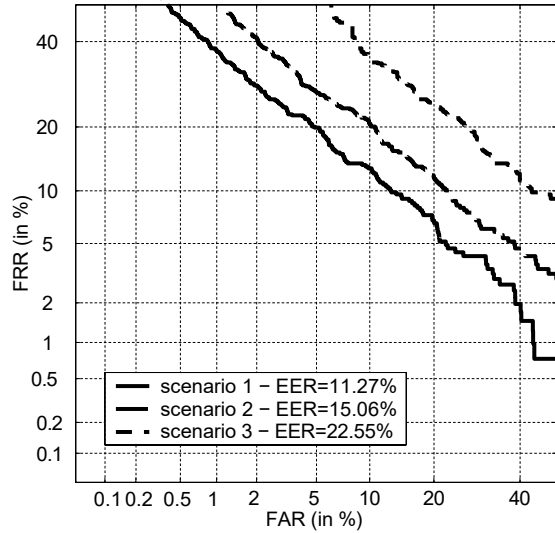 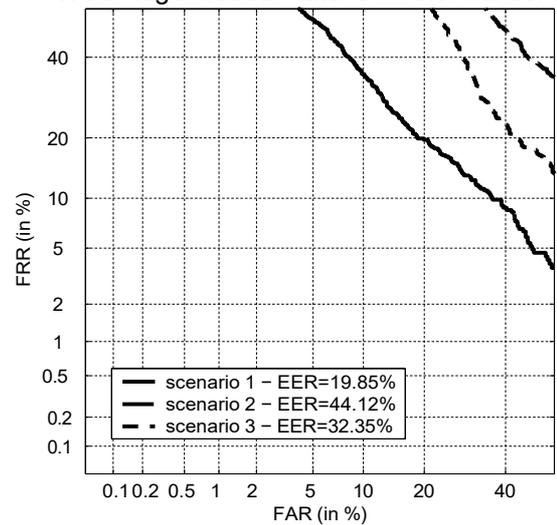

**Figure 8. DET curves for the ATVS system using the optical sensor (left) and the thermal sensor (right).**

**Thermal sensor**. Again a drop in the system performance is observed as compared to the NIST software. Worth noting that in this case the increase of the EER for the scenario 1 from the optical to the thermal sensors is not as significant as in the NIST case. This may be explained because the ridge-based system is robust to low quality samples [12].

In summary, the overall performance of the ridge-based system is worse than the minutiae-based one, however, it is less vulnerable to direct attacks and it is also more robust to low quality samples.

## 5  Conclusions

A new method to generate gummy fingers was presented. A medium-size fake fingerprint database was described and two different fingerprint verification systems, one minutiae-based and one ridge-based, were evaluated on it. Three different scenarios were considered in the experiments, namely: *i*) enrollment and test with real fingerprints, *ii*) enrollment and test with fake fingerprints, and *iii*) enrollment with real fingerprints and test with fake fingerprints. Results for an optical and a thermal sweeping sensors were given.

The NIST fingerprint verification system performs better with the optical sensor but it is also more vulnerable to direct attacks on it as compared to experiments on the thermal sensor.

The overall performance of the ridge-based system was worse than the minutiae-based one, however, it showed to be less vulnerable to direct attacks and it was also more resistant to low quality samples.

## Acknowledgments

This work has been supported by Spanish Ministry of Defense, BioSecure NoE and the TIC2003-08382-C05-01 project of the Spanish Ministry of Science and Technology. J. G.-H. is supported by a FPU Fellowship from the Ministerio de Educacion y Ciencia (Spanish Ministry of Science). J. F.-A. and F. A-F are supported by a FPI Fellowship from Comunidad de Madrid.